\documentclass[letterpaper, 10 pt, conference]{ieeeconf}
\IEEEoverridecommandlockouts

\usepackage{amsmath}
\usepackage{gensymb}
\usepackage[table]{xcolor}
\usepackage{cite}
\usepackage[colorlinks=true, citecolor=cyan]{hyperref}
\usepackage{booktabs}
\usepackage{graphicx}
\usepackage{multirow}
\usepackage{wrapfig}
\usepackage{longtable}
\usepackage{mdframed}

\title{\LARGE \bf
Where Should I Join? Robot Group Joining via \\ Language-Guided Goal Prediction
}

\author{Zilin Fang$^1$, Zishuo Wang$^1$, Gim Hee Lee$^{1}$, and David Hsu$^{1,2}$
\thanks{$^{1}$School of Computing, $^{2}$Smart Systems Institute, National University of Singapore, Singapore. Correspond to
        {\tt\small zilin.fang@u.nus.edu 
        }}%
}

\begin{document}

\maketitle
\thispagestyle{empty}
\pagestyle{empty}

\begin{abstract}

Social navigation typically assumes a specified goal and focuses on reaching it while respecting social conventions, whereas robot group joining requires predicting where to join based on the group's real-time activity and formation. This is a highly semantic task, yet an important capability for applications such as robotic guide dogs and autonomous mobility scooters. We formulate language-grounded robot group joining: given an observation and a natural-language description of a target group, the robot identifies the relevant group members and predicts socially compliant joining poses. For grounding, we generate structured candidate subsets through recursive spectral partitioning and rank them with a language-conditioned image--geometry model. Given the grounded group, a goal predictor leverages human-formation priors to produce a multimodal energy--orientation map over feasible robot poses. Experiments on conversations, queues, and audiences across varying group sizes, crowd densities, and visual ambiguities show that our method achieves competitive grounding accuracy with sub-second inference and outperforms all baselines in joining-pose prediction. Real-robot experiments further demonstrate group joining in both static and dynamically changing interactions. Project webpage: \href{https://robot-join.github.io}{\textcolor{violet}{robot-join.github.io}}

\end{abstract}

\section{Introduction}
Imagine traveling through a downtown square on an autonomous scooter and deciding to grab a coffee to go. When you arrive at the stall, a queue has already formed. Rather than teleoperating the vehicle or joining the line yourself, you simply say: ``Join the queue at the coffee stall,'' and let the vehicle handle the rest. This scenario highlights a challenge beyond typical social navigation: the goal is not pre-specified but must be inferred on-the-fly from the activity and configuration of a human group. The appropriate joining pose depends on who belongs to the group, where they are, and how they are oriented, all determined at runtime from a cluttered scene. A robot must therefore ground the natural language instruction to the correct group and predict a joining pose that is geometrically feasible and socially compliant.

\begin{figure}[t]
  \centering
  \includegraphics[width=0.91\linewidth]{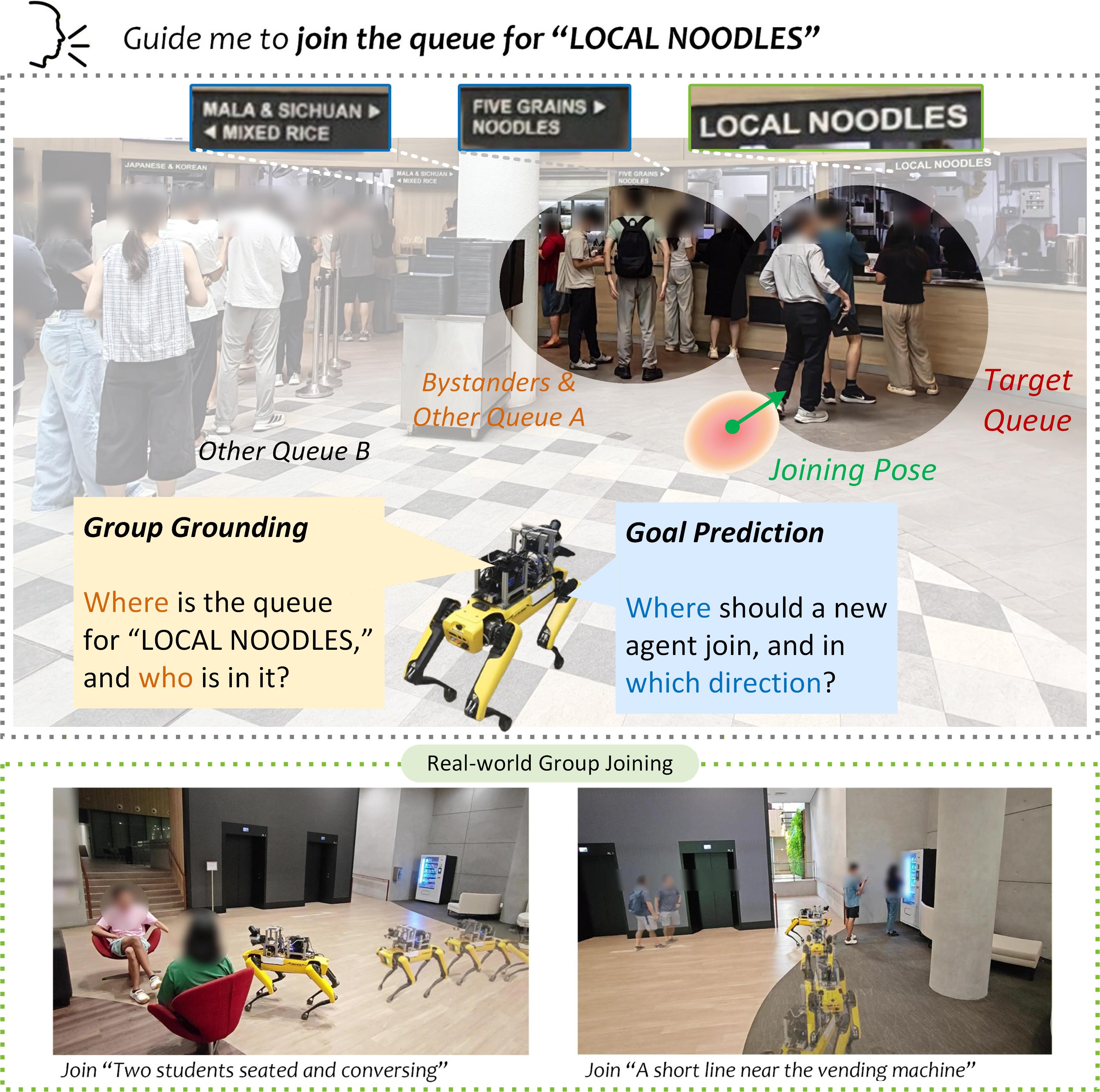}
  \vspace{-0.8em}
  \caption{Robot group joining from language instructions. Given a target group description, our method identifies the members and predicts socially compliant, geometrically valid joining poses.}
  \label{fig:teaser}
  \vspace{-2em}
\end{figure}

Most social navigation methods assume that a navigation goal is already available~\cite{hirose2023sacson, narasimhan2025olivia, fang2026socialnav} and focus on reaching it while respecting nearby people. Group detection methods instead reason about which observed people belong together~\cite{taylor2020robot, wang2022group}, but do not determine a task-specific navigation target. Neither directly addresses scenarios where the goal itself emerges from a language-specified social interaction. We therefore formally define the robot group joining problem as a language-grounded goal prediction task (Fig.~\ref{fig:teaser}): given an RGB-D observation of the current scene and a natural language instruction specifying the target group, the robot predicts a distribution over joining poses that depends on the target group's membership, geometry, and orientation. We decompose the task into a two-stage pipeline. First, \textit{group grounding} identifies the task-relevant subset of observed people corresponding to the language instruction. We generate structured candidate subsets through divisive spectral clustering and rank them with a model trained via pairwise comparison. This combines visual semantics with the spatial coherence of the hypothesized members while avoiding exhaustive enumeration of all possible subsets. The grounded group is then mapped into a top-down representation encoding group membership and human poses. Second, \textit{goal prediction} uses this representation to generate an energy--orientation map over valid joining poses. The predictor is trained on simulated human formations derived from three recurring attention topologies: sequential attention, mutual attention, and directed shared attention.

We evaluate our method on diverse scenes featuring these three fundamental human group topologies: conversations, queues, and audiences, spanning various group sizes, crowd densities, and visual ambiguities. Our grounding stage achieves competitive agreement with human annotations while operating significantly faster than large Vision-Language Model (VLM) baselines. The proposed pipeline outperforms the evaluated heuristic and VLM baselines in goal prediction, while real-robot experiments show reliable group joining compared with Vision-Language-Action (VLA) navigation baselines. The system successfully joins both static and dynamic groups with collision-free navigation and efficient goal updates, highlighting its potential for assistive applications such as robotic guide dogs and autonomous wheelchairs.

\section{Related Work}

\subsection{Social Navigation and Group Detection}
Social navigation methods learn human-aware policies from social rewards~\cite{hirose2023sacson,narasimhan2025olivia}, demonstrations~\cite{karnan2022socially}, or foundation-model priors~\cite{song2024vlm,fang2026socialnav} but generally assume a specified goal. SANG~\cite{schmuck2025sang} explicitly incorporates group structure into navigation rewards using SALSA~\cite{alameda2015salsa} while still navigating toward a predefined position. Group joining additionally requires estimating the goal itself from the target group's configuration and activity. This requirement is closely related to human group perception. Human group detection has been extensively studied in images and robot-centric scenes~\cite{choi2014discovering,martin2021jrdb,taylor2020robot}, with recent approaches modeling interpersonal relations through graph representations~\cite{li2022self,jahangard2023real,yokoyama2025dynamic}. These methods estimate group membership among observed people rather than grounding a language-specified target group. The closest navigation-related work~\cite{eirale2024learning} learns social costs for queue joining; yet it also assumes a predefined goal coordinate and is demonstrated on structured, straight queues. In contrast, we infer an open-ended joining pose directly from the observed formation and a language instruction.

\subsection{F-formation Detection and Group Joining}
F-formation detection has been studied from overhead~\cite{setti2015f, hedayati2020reform, raman2022conflab} and egocentric view~\cite{barua2020let, pathi2022detecting} images for years. Beyond detection, prior work predicts positions of undetected people~\cite{hedayati2022predict} or suitable robot positions in conversational groups~\cite{barua2020let, hedayati2023should}. AGIR~\cite{pathi2022detecting} samples robot positions for evaluating F-formation detection across viewpoints, rather than for predicting joining locations. These methods mainly target conversations and typically assume that the relevant interaction has already been identified. While the taxonomy in~\cite{barua2024enabling} shows that social interactions exhibit structured spatial patterns via F-formations, it does not address how to actively join diverse functional groups, where a valid joining position should avoid obstructing existing members or disrupting their activities and, when applicable, preserve the existing formation structure. Addressing this gap, we abstract these formation transitions into core attention topologies and train a model to predict joining poses across various group types.

\subsection{Language grounding for navigation}
Foundation models are increasingly applied to ground free-form language instructions to navigation-relevant targets. For example, LM-Nav~\cite{shah2023lm} and HOV-SG~\cite{werby2024hierarchical} associate language with objects and landmarks, enabling long-horizon navigation without fine-tuning. For human targets, SOCRATES~\cite{park2023socrates} employs a chained language-model pipeline to find and approach a target individual described in free-form text, while text-based person re-identification~\cite{jiang2025attributes} addresses cross-modal retrieval of persons from natural language. Many works also translate language instructions directly into navigation actions; NaVILA~\cite{cheng2024navila} generates mid-level language-form action primitives, and NavMorph~\cite{yao2025navmorph} predicts stepwise positional displacements. These approaches primarily target landmarks, objects, and terrain-based navigation, but do not extend language grounding to socially defined groups or predict valid joining poses once a group is identified. Our work bridges these threads by grounding language group descriptions to in-group individuals and then predicting geometrically and socially valid joining poses.

\section{Task Formulation}
\label{sec:formulate}
\begin{figure}[t]
  \centering
  \includegraphics[width=0.98\linewidth]{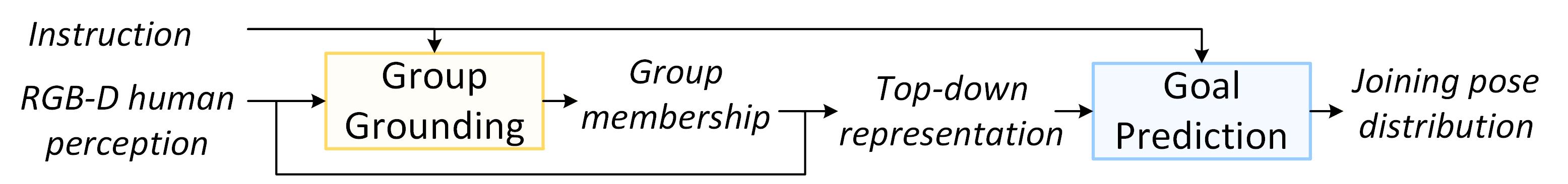}
  \vspace{-0.9em}
  \caption{Overview of the language-grounded goal prediction task.}
  \vspace{-1.8em}
  \label{fig:task_pipeline}
\end{figure}

We consider a mobile robot joining a socially interacting human group from a single RGB-D observation. Let $O$ denote the paired RGB image and depth map, and let $T$ be a natural-language instruction that unambiguously specifies the intended group, e.g., join ``the conversation with the woman holding a bag.'' The desired output is a geometrically valid and socially compliant target-pose distribution $p_{\text{target}}(x, y, \theta \mid O, T)$, where $(x,y)$ is the robot position on the ground plane and $\theta$ its heading. We model a distribution rather than a single waypoint because socially acceptable joining behavior is inherently multimodal. This prediction problem is addressed with two sequential stages (Fig.~\ref{fig:task_pipeline}): (i) Group grounding, which identifies a task-relevant subset of detected persons that is semantically consistent with the group described by $T$, and (ii) Goal prediction, which estimates a target pose distribution conditioned on the grounded group members.

\begin{figure}[tbhp]
  \centering
  \includegraphics[width=0.95\linewidth]{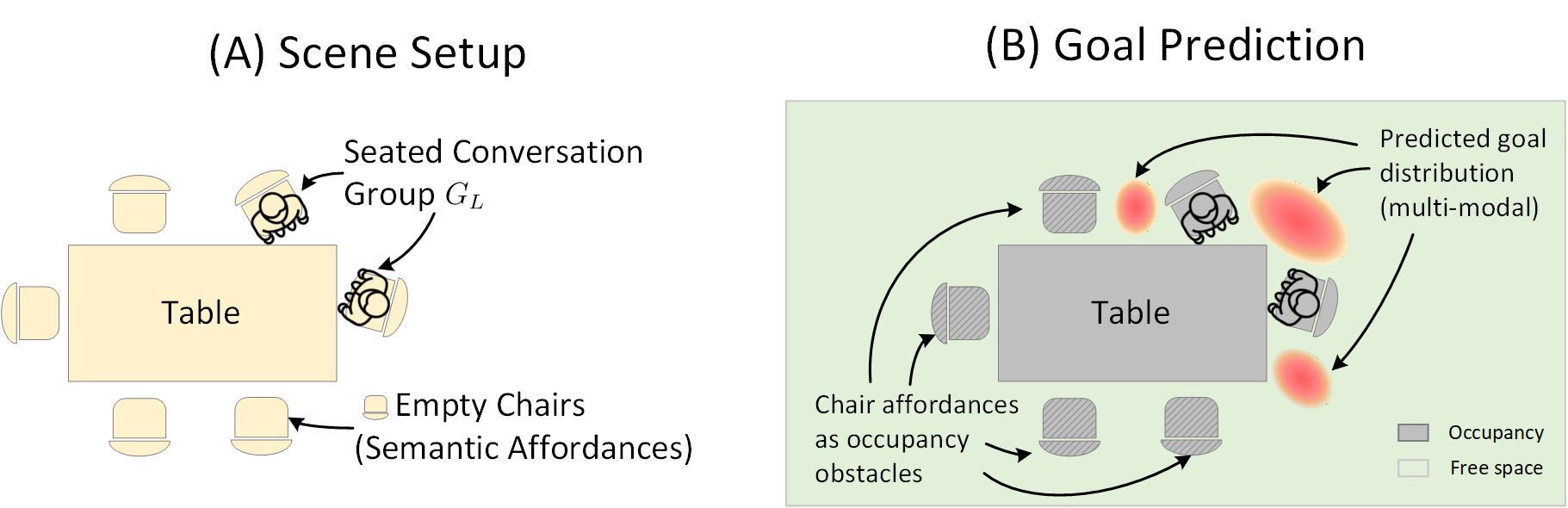}
  \vspace{-0.8em}
  \caption{Illustration of our assumption. Environmental constraints restrict infeasible regions rather than prescribing target positions.}
  \vspace{-1.1em}
  \label{fig:assumption}
\end{figure}

\begin{figure*}[]
  \centering
  \includegraphics[width=0.93\linewidth]{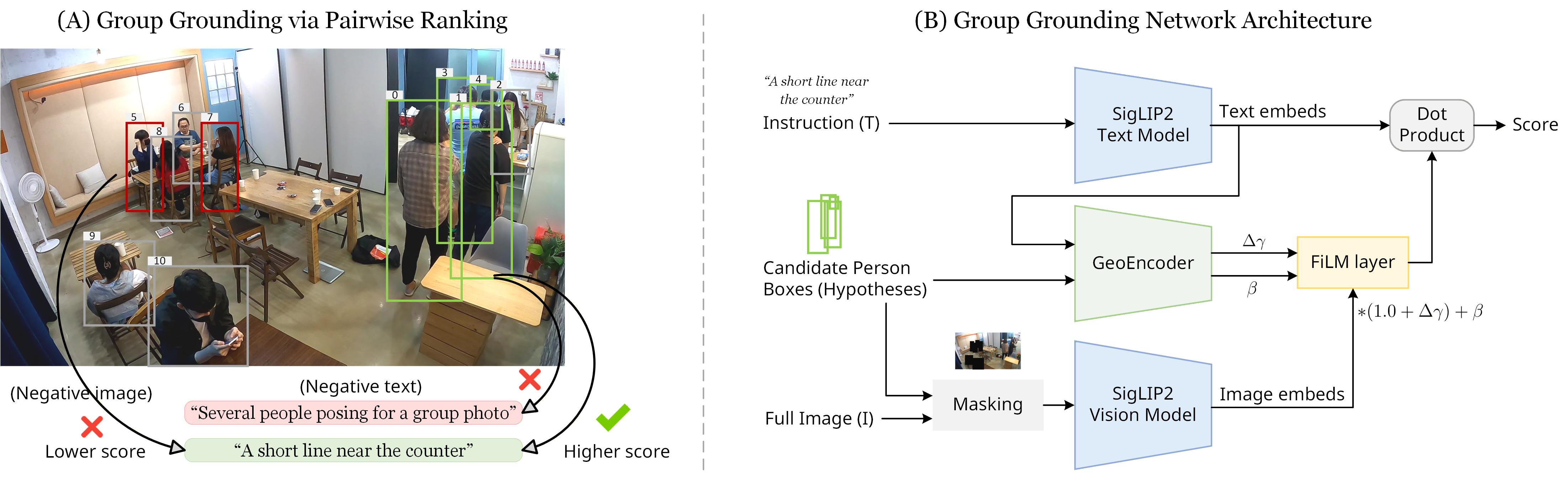}
  \vspace{-1.2em}
  \caption{Language-guided group grounding. (A) The grounding module is trained using a pairwise ranking mechanism. The network (B) evaluates multiple hypotheses of candidate person bounding boxes, scoring each against the natural language instruction to identify the target group.}
  \label{fig:ground}
  \vspace{-1.8em}
\end{figure*}

\textbf{Group Grounding.} Let $G = \{g_0, g_1, \ldots, g_{N-1} \}$ denote the $N$ persons returned by a human detector. Each person $g_i$ contains a unique ID and an image-space bounding box. Group grounding produces a subset $G_T \subseteq G$ that best corresponds to the group described by $T$. Here, correspondence is task-oriented: it requires sufficient geometric information for determining a valid joining pose rather than exhaustive recovery of all members of the semantic group. For spatially dispersed groups, for example, a locally coherent subset may still be sufficient if it preserves the interaction context necessary for joining. The surrounding environment serves as a semantic anchor, informing the functional purpose and spatial extent of a group (e.g., furniture arrangement, room context), but is retained in the original observation space without explicit structural extraction. The output is then transformed into an intermediate top-down representation encoding group membership and estimated human poses.

\textbf{Goal Prediction.} Given $G_T$, goal prediction estimates $p(x, y, \theta \mid G_T, O, T)$. We assume environmental elements act as occupancy obstacles; thus, valid joining positions are primarily determined by the group's spatial arrangement and mutual attention. The environment restricts infeasible regions but does not explicitly prescribe target positions. Take Fig.~\ref{fig:assumption} as an example. Two people are conversing while seated side-by-side at a table. A person joining could take an adjacent chair, sit across the table, or stand nearby; all are socially valid. Our model predicts poses in free space that satisfy spatial and social constraints of the group, without committing to furniture-specific positions. This is consistent with the inherently multi-modal nature of joining behavior, where a single correct answer rarely exists.

\begin{figure*}[]
  \centering
  \includegraphics[width=0.95\linewidth]{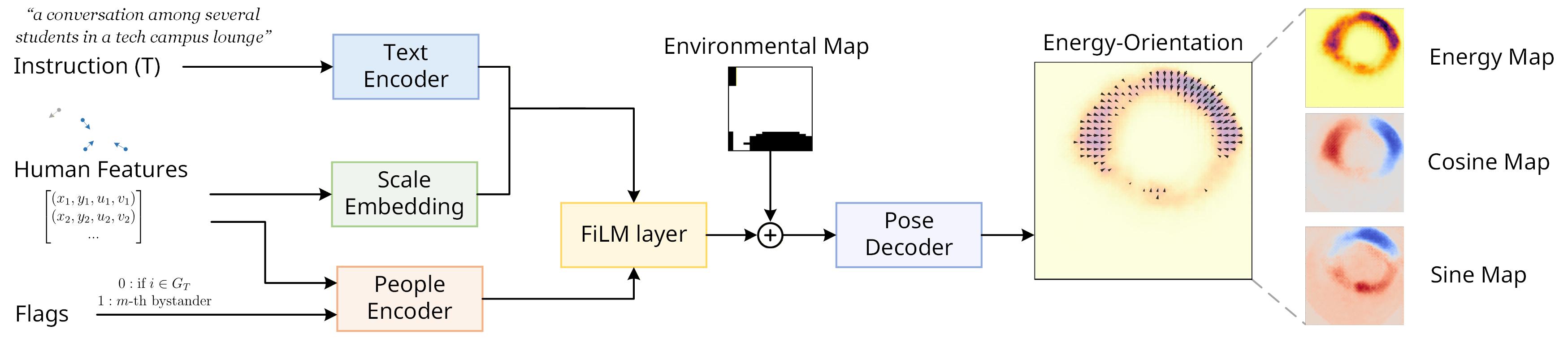}
  \vspace{-1.2em}
  \caption{Goal prediction network architecture. Conditioned on language instructions and identified group members, the network predicts a joining-pose distribution using top-down human representations, outputting a 3-channel map of positional energy and sine/cosine orientations.}
  \label{fig:goal_pred}
  \vspace{-1.8em}
\end{figure*}

\section{Group Grounding}
\label{sec:group_ground}
Rather than directly predicting $G_T$, we first generate a compact set of candidate group hypotheses $\mathcal{H}$ and then score each hypothesis against language instruction $T$. Each hypothesis $H \in \mathcal{H}$ is a subset of detected persons, represented by their visual observations and spatial relationships. The highest-scoring hypothesis is selected as $G_T$. We assume that each group has a predominant identifiable activity that defines the group at the scene level, despite possible local sub-interactions (e.g., brief conversations within a queue) among its members.

\subsection{Hypothesis Generation}
Exhaustively considering all possible person subsets is impractical: an $N$-person scene yields up to $|\mathcal{H}_{all}| = \sum_{k=2}^{N} \binom{N}{k} = 2^N - 1 - N$ subsets (excluding singletons and the empty set), growing rapidly for dense scenes and producing many redundant hypotheses that waste ranking capacity. We therefore generate a compact and structured hypothesis set $\mathcal{H}$ via graph partitioning, based on the observation that group members are typically spatially proximate.

We adopt spectral clustering with local scaling~\cite{zelnik2004self}. Human detections are projected onto the top-down 2D plane using depth and camera intrinsics, and pairwise affinities are computed with self-tuning local scales estimated from each person's k-th-neighbor distance. The normalized random-walk Laplacian is then computed with $L = I - D^{-1}A, D_{ii} = \sum_j A_{ij}$, where $I$ is the identity matrix and $A$ the affinity matrix. Rather than applying a single zero-crossing cut on the Fiedler vector (the eigenvector associated with the second-smallest eigenvalue of $L$), we sweep multiple thresholds across its distribution. Specifically, we consider thresholds between the $10$th and $90$th percentiles, divided into $6$ evenly spaced intervals as an accuracy–efficiency trade-off. Each threshold induces a different binary partition of the current node set, yielding multiple candidate sub-graphs from a single eigenvector. This strategy naturally extends spectral clustering by increasing partition diversity and is motivated by two considerations: (i) social group boundaries are often ambiguous (e.g., unevenly spaced queue members or nearby bystanders), where no single geometric threshold is reliably optimal; and (ii) it improves robustness to noise in depth estimation and camera calibration.

Partitioning is applied recursively in a top-down manner. Starting from the full person set $G$, each node set is split into two subsets, which are further partitioned until singleton subsets are reached. All threshold-swept sub-graphs from every recursion level are collected as candidate hypotheses. The resulting hypothesis set $\mathcal{H}$ spans different group sizes and boundary granularities, and is finally passed to the ranking model to identify the target group $G_T$.

\subsection{Hypothesis Ranking}
To evaluate the candidate set $\mathcal{H}$, we learn to score each hypothesis
according to its compatibility with instruction $T$. We fine-tune SigLIP2~\cite{tschannen2025siglip} with a pairwise ranking objective, as illustrated in Fig.~\ref{fig:ground} (A). Each sample consists of a (\textit{text, image, candidate subset}) triple, where persons outside the candidate subset are masked so that the visual input focuses on the hypothesized group members. To incorporate their spatial relationships, a GeoEncoder applies person–person self-attention and person–language cross-attention. The resulting geometric features are converted into modulation parameters $(\Delta\gamma, \beta)$, which adapt the image embeddings with task-relevant spatial context. Finally, the fused image–geometry representation is compared with the text embedding to produce a similarity score, shown in Fig.~\ref{fig:ground} (B).

The model is trained with two group datasets: Group Discovery~\cite{choi2014discovering} and Café~\cite{kim2024towards}. To reduce dataset imbalance and avoid majority-class bias, approximately $8$k overrepresented Café samples are removed. We also discard posed close-up images from Group Discovery, as they do not represent natural social interactions. After filtering, $1{,}421$ images remain. Since no captions are provided, we generate visually grounded interaction descriptions using Qwen3.5-397B-A17B~\cite{qwen3.5}, with multiple variants emphasizing different social and visual attributes. Negative text samples are created as mismatched activity descriptions for the same group. Negative image samples are generated by perturbing ground-truth groups $G_T^*$ using four strategies: (i) removing members, (ii) adding non-members, (iii) swapping members, and (iv) random sampling. A pair $(G^+, G^-)$ is kept only if their F1 difference exceeds $0.05$ (relative to $G_T^*$ with F1 = $1.0$), ensuring meaningful difficulty. This yields a mix of easy and hard training pairs. The final dataset contains $10{,}624$ training pairs and $1{,}000$ validation pairs. We optimize a dynamic margin ranking loss, where the margin for image-based pairs scales with the F1 gap ($\times 0.3$), encouraging stronger separation for larger membership errors, while text-based pairs use a fixed $0.2$ margin. At inference, we evaluate all candidate hypotheses in a batched forward pass and select the highest-scoring subset as the final grounded group.

\section{Goal Prediction}
\label{sec:goal_pred}
Given the grounded group members $G_T$, we predict a distribution over plausible robot joining poses. Socially valid joining follows patterns from empirical studies of human F-formations, which describe how group spatial structure adapts when new members join~\cite{barua2024enabling}. We leverage this taxonomy to generate structured training data and learn a model that predicts $p(x, y, \theta \mid G_T, O, T)$ as an energy--orientation map. The three-channel output consists of a spatial energy map $E(x,y)$, where each location $(x, y)$ is assigned a scalar energy (lower indicates higher likelihood), and two orientation maps encoding the cosine and sine of the robot heading. The architecture is illustrated in Fig.~\ref{fig:goal_pred}.

We construct a top-down representation from the grounded group and detected humans. Each person is described by $(x_i,y_i,u_i,v_i)$, where $(x_i,y_i)$ denotes position and $(u_i,v_i)$ the orientation vector, together with a binary indicator of membership in $G_T$; nearby bystanders are retained for local context. Positions are normalized by the global scale and encoded by a People Encoder, with group membership incorporated through additive embeddings. The global scale is encoded separately and concatenated with the language instruction to condition the person features through a FiLM layer. The resulting features are reshaped and passed to a transposed-convolutional Pose Decoder, with a scene occupancy map constraining predictions to free space. At inference, the location with minimum energy in $E(x,y)$ gives the target position $(x,y)$, while the corresponding cosine and sine values define the target heading $\theta$. The orientation field overlaid on the energy map is also visualized in Fig.~\ref{fig:goal_pred}.

To train this module, we simulate group configurations based on the $29$ F-formation transitions in~\cite{barua2024enabling}, abstracting them by their spatial and attentional structure, i.e., the relationship between participant positions, orientations, and their shared locus of attention, into three fundamental \textit{attention topologies}: i) \textbf{sequential attention}, where individuals follow one another with a shared global direction (e.g., column, diagonal); ii) \textbf{mutual attention}, where participants orient toward a shared o-space, with formations such as triangles, horseshoes, semicircles, etc. viewed as discrete samplings of an underlying continuous circular o-space; and iii) \textbf{directed shared attention}, where participants jointly attend to an external focal entity (e.g., a speaker or exhibit).

For simulation, sequential groups are generated by sampling a direction uniformly over $360\degree$ and placing members along the corresponding line. Mutual-attention groups contain $2$--$9$ members sampled along an arc of a randomly sized circle and oriented toward its center. Directed-shared-attention groups are arranged on one or more arcs or rows facing a sampled external focal point, with variable spacing and angular density. We further perturb human positions and orientations, add random bystanders, and synthesize occupancy maps with randomly placed geometric obstacles that avoid person locations. Each scene is paired with a topology-specific language instruction using four template variants per topology. Instruction components are sampled from pools of $60$ location phrases, $30$ subject-group descriptions, and $30$ topology-appropriate action verbs (e.g., buy tickets, order lunch, debate, joke, or watch a demo). The resulting dataset contains $10{,}000$ training and $2{,}000$ validation samples.

\begin{table}[]
    \centering
    \setlength{\tabcolsep}{4pt}
    \caption{Comparison of agreement-weighted grounding performance}
    \vspace{-1.1em}
    \label{tab:ground_result}
    \resizebox{\linewidth}{!}{%
    \begin{tabular}{@{}lcccc@{}}
    \toprule
        Method & Precision & Recall & F1 Score & Time (s) (ground.) \\
    \midrule
        \multicolumn{5}{l}{\textit{Zero-shot prompting}} \\
        Molmo2~\cite{clark2026molmo2}             & 0.4274 & 0.6096 & 0.4724 & 1.03 \\
        Qwen3.5-397B-A17B (w/o t.)~\cite{qwen3.5} & 0.7981 & 0.8242 & 0.7885 & 2.41 \\
        Qwen3.5-397B-A17B (t.)~\cite{qwen3.5}     & 0.7934 & 0.8471 & 0.8027 & 5.31 \\
        Gemini 3.1 Pro~\cite{geminipro}  & \textbf{0.8378} & 0.8342 & 0.8176 & 8.73 \\
        Gemini 3 Flash~\cite{geminiflash3} & 0.7916  & 0.8809 & \underline{0.8240} & 8.63 \\
        Gemini 3.5 Flash Lite~\cite{geminiflash35} & 0.7938  & 0.8382 & 0.7932 & 2.60 \\
        GPT-5.5~\cite{gpt55}                      & 0.8059 & 0.8476 & 0.8178 & 3.79 \\
    \midrule
        \multicolumn{5}{l}{\textit{Few-shot in-context learning}} \\
        Gemini 3 Flash~\cite{geminiflash3}         & 0.7794 & \textbf{0.8987} & 0.8208 & 4.19 \\
        GPT-5.5~\cite{gpt55}                      & 0.8022 & \underline{0.8825} & \textbf{0.8311} & 5.48 \\
    \midrule
        VLM-GroupnessGraph~\cite{yokoyama2025dynamic}* & 0.4313 & -- & 0.5736 & -- \\
        \textbf{Ours} & \underline{0.8171} & 0.8393 & 0.8125 & \textbf{0.936} \\
    \bottomrule
    \end{tabular}%
    }
    \parbox{\linewidth}{\scriptsize *Limited to group detection without grounding, despite the same training source~\cite{kim2024towards}.}
    \vspace{-2em}
\end{table}

\begin{table}[]
    \centering
    \setlength{\tabcolsep}{4pt}
    \caption{Comparison of goal prediction performance}
    \vspace{-1.1em}
    \label{tab:pose_result}
    \resizebox{\linewidth}{!}{%
    \begin{tabular}{@{}l|cccc@{}}
    \toprule
        \multirow{2}{*}{Model} & Position  & Orientation & Pose & Inference \\ 
        & Validity (\%) & Validity (\%) & Validity (\%) & Time (s) \\
    \midrule
        \multicolumn{5}{l}{\textit{Zero-shot prompting}} \\
        Molmo2\cite{clark2026molmo2} & 14.05$\pm$4.55          & 40.95$\pm$4.02           & 10.71$\pm$5.86          & 1.22               \\
        Qwen3.5-397B-A17B (w/o t.)~\cite{qwen3.5} & 30.48$\pm$5.92          & 45.24$\pm$4.49           & 22.62$\pm$3.41          & 3.39               \\
        Qwen3.5-397B-A17B (t.)~\cite{qwen3.5}     & 45.95$\pm$5.73          & 70.71$\pm$5.84           & 39.52$\pm$6.47          & 28.70              \\
        Gemini 3.1 Pro~\cite{geminipro}   & 54.05$\pm$3.90          & 78.57$\pm$4.76           & 46.67$\pm$4.08      & 27.94              \\
        Gemini 3 Flash~\cite{geminiflash3} & 52.38$\pm$6.73          & 87.86$\pm$2.08           & 51.19$\pm$6.27      & 25.50              \\
        Gemini 3.5 Flash Lite~\cite{geminiflash35} & 50.71$\pm$6.35 & 68.81$\pm$3.06           & 36.19$\pm$5.36      & 3.78              \\
        GPT-5.5~\cite{gpt55}             & 50.24$\pm$3.80          & 83.33$\pm$3.55           & 45.95$\pm$2.26      & 38.17              \\
    \midrule
        \multicolumn{5}{l}{\textit{Few-shot in-context learning}} \\
        Gemini 3 Flash~\cite{geminiflash3} & 56.67$\pm$9.31          & 88.57$\pm$4.32          & 53.57$\pm$7.29      & 27.80 \\
        GPT-5.5~\cite{gpt55}               & 61.19$\pm$8.98          & 91.43$\pm$2.30          & 59.76$\pm$8.73      & 49.25 \\
    \midrule
        \textbf{Ours}           & \textbf{90.47$\pm$3.55} & \textbf{96.67$\pm$2.30}  & \textbf{89.76$\pm$4.21}  & \textbf{1.108}    \\
    \midrule
    \midrule
        Oracle-group + handcrafted rule~\cite{barua2024enabling}    & 71.43        & 83.33        & 64.29      & -- \\
        Oracle-group + multivariate reg.~\cite{hedayati2023should}  & 38.10        & 78.57        & 35.71      & -- \\
        \textbf{Ours (oracle-group)} & \textbf{95.71$\pm$2.70}  & \textbf{97.62$\pm$2.97}  & \textbf{94.76$\pm$3.33}      & -- \\
    \bottomrule
    \end{tabular}%
    }
    \parbox{\linewidth}{\scriptsize GPT-5.5 Pro is excluded because it requires over 100 seconds per query.}
    \vspace{-3em}
\end{table}

\begin{table*}[]
    \centering
    \setlength{\tabcolsep}{2pt}
    \caption{Real-robot experiment results}
    \vspace{-1.1em}
    \label{tab:robot_result}
    \resizebox{\textwidth}{!}{%
    \begin{tabular}{@{}l|cccc|cccc|cccc|cccc@{}}
    \toprule
        & \multicolumn{4}{c|}{\textbf{Seated Conversation}}
        & \multicolumn{4}{c|}{\textbf{Standing Conversation}}
        & \multicolumn{4}{c|}{\textbf{Static Queue}}
        & \multicolumn{4}{c}{\textbf{Dynamic Queue}} \\
    \cmidrule(lr){2-5}
    \cmidrule(lr){6-9}
    \cmidrule(lr){10-13}
    \cmidrule(lr){14-17}
        Method 
        & SR & CR & NT (s) & PSV (s)
        & SR & CR & NT (s) & PSV (s)
        & SR & CR & NT (s) & PSV (s)
        & SR & CR & NT (s) & PSV (s) \\
    \midrule
    NaVILA~\cite{cheng2024navila}
        & 0/5 & 2/5 & 54.07 & 0
        & 0/5 & 0/5 & --[stalls] & --
        & 0/5 & 5/5 & --[stalls] & --
        & 0/5 & 0/5 & --[stalls] & -- \\
    OmniVLA~\cite{hirose2025omnivla}
        & 0/5 & 0/5 & 36.03 & 0
        & 0/5 & 0/5 & 48.71 & 0
        & 2/5 & 2/5 & 42.75 & 8.93
        & 1/5 & 3/5 & 55.46 & 10.89 \\
    Ours (ground.) w. sample-select
        & 4/5 & 1/5 & 22.08 {\tiny(+37.2)}  & 3.51
        & 3/5 & 0/5 & 25.57 {\tiny(+40.67)} & 3.96
        & 4/5 & 1/5 & 27.01 {\tiny(+17.35)} & \textbf{0}
        & 2/5 & 1/5 & 25.43 {\tiny(+15.01)} & 0 \\
    \rowcolor{gray!20} \textbf{Ours}
        & \textbf{5/5} & \textbf{0/5} & \textbf{20.63} & \textbf{0}
        & \textbf{5/5} & \textbf{0/5} & \textbf{24.01} & \textbf{0}
        & \textbf{4/5} & \textbf{0/5} & \textbf{26.31} & 0.88
        & \textbf{4/5} & \textbf{0/5} & \textbf{28.17} & \textbf{0} \\

    \bottomrule
    \end{tabular}%
    }
    \vspace{-1.8em}
\end{table*}

\begin{figure*}[]
  \centering
  \includegraphics[width=\linewidth]{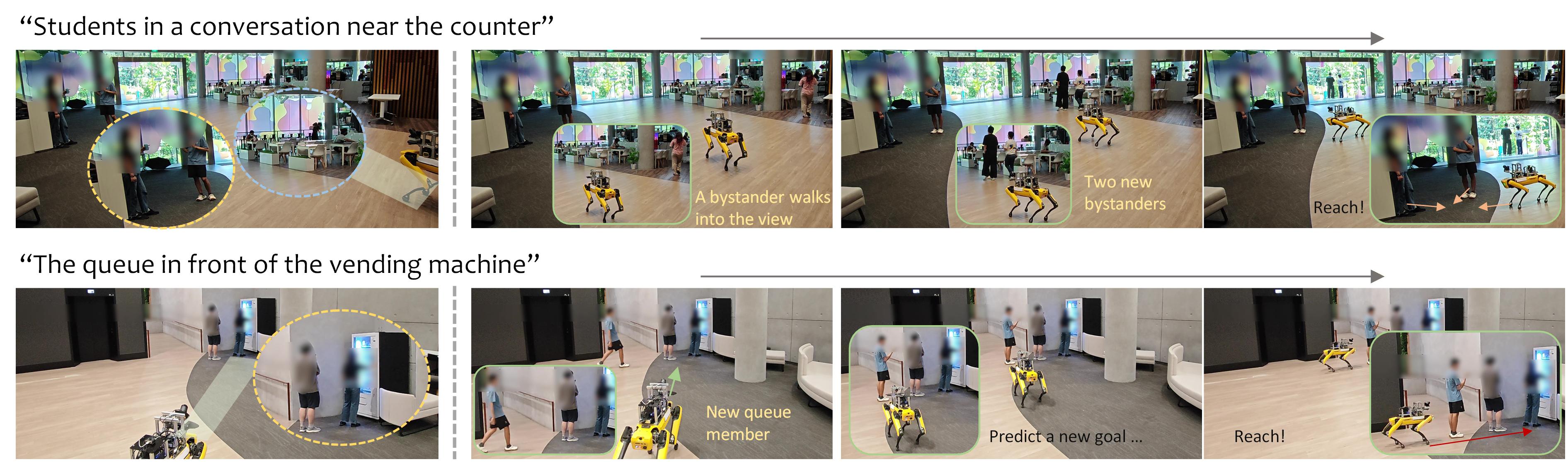}
  \vspace{-2.2em}
  \caption{Snapshots of a robot joining a standing conversation group and a dynamic queue.}
  \label{fig:robot_exp}
  \vspace{-1.5em}
\end{figure*}

\section{Experiments}
\label{sec:experiment}

\subsection{Experimental Setup}
As no prior work addresses the complete pipeline from a language instruction to a goal pose conditioned on in-scene human activity, we evaluate the two stages both independently and end-to-end: (i) group grounding, (ii) goal prediction given the ground-truth group, and (iii) the full pipeline using predicted groups. The \textbf{offline benchmark} contains three representative interaction scenarios corresponding to the attention topologies defined in Sec.~\ref{sec:goal_pred}: \textit{queue} for sequential attention, \textit{conversation} for mutual attention, and \textit{audience} for directed shared attention. Scenarios are drawn from the egocentric robot datasets JRDB~\cite{martin2021jrdb} and SCAND~\cite{karnan2022socially}, supplemented by hand-held camera footage. The benchmark comprises $15$ conversation, $15$ queue, and $12$ audience scenarios ($42$ total); $11$ audience scenes were manually collected at gallery exhibitions and street performances because such interactions are under-represented in existing datasets. The benchmark spans diverse crowd densities, group sizes, configurations, and grounding ambiguities. \textbf{Real-robot experiments} include seated and standing conversations and static and dynamic queues, with the dynamic queue containing an additional person who joins during navigation.

\begin{figure*}[]
  \centering
  \includegraphics[width=0.95\linewidth]{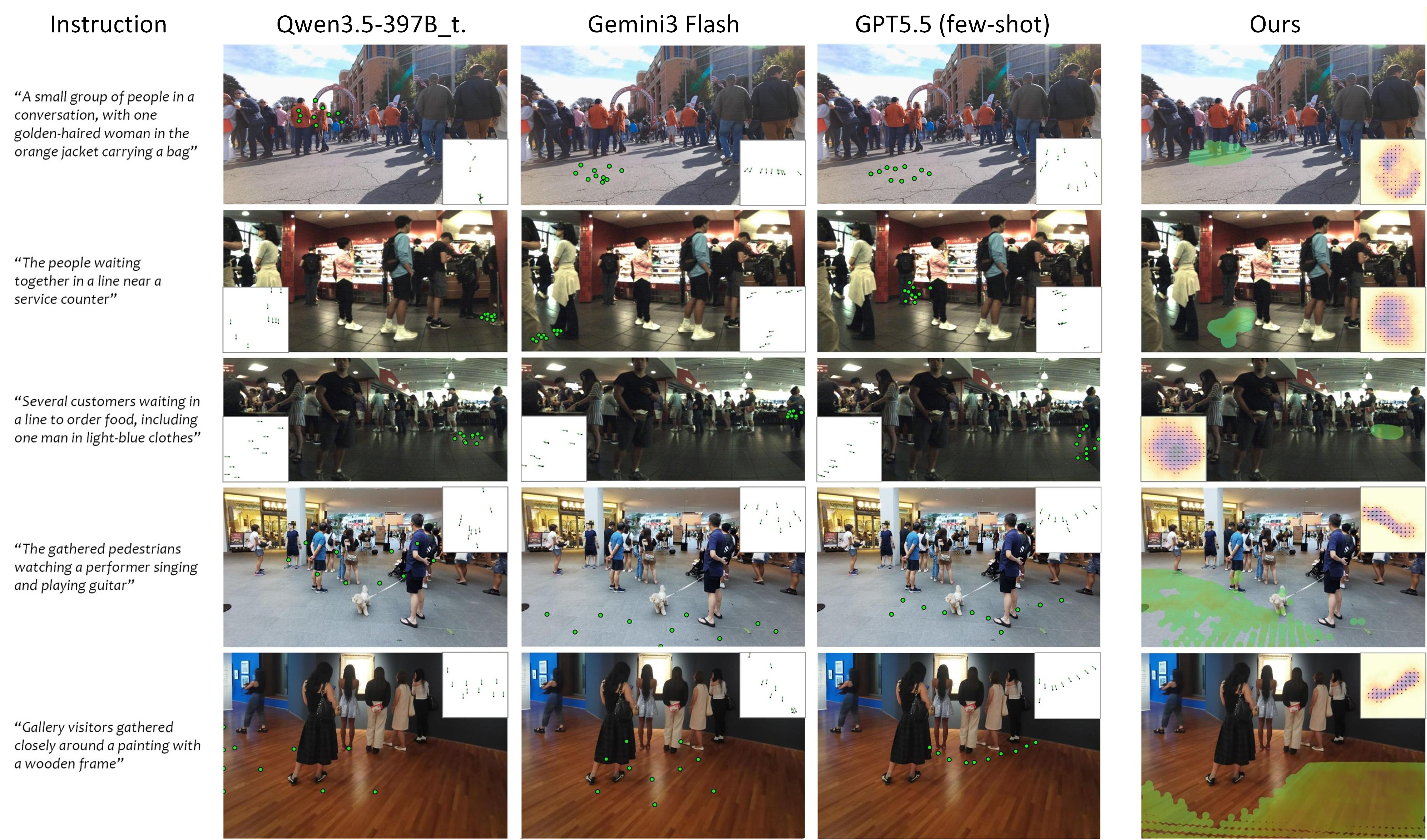}
  \vspace{-0.9em}
  \caption{Visualization of goal prediction results on offline scenarios. Orientations are displayed in small square insets for each scene. Green dots denote VLM predictions, with boundary points indicating predictions outside the image frame.}
  \label{fig:exp_result}
  \vspace{-1.2em}
\end{figure*}

\textbf{VLM Baselines.} 
For group grounding and end-to-end offline evaluation, we compare against several VLMs: Molmo 2~\cite{clark2026molmo2}, Qwen3.5-397B-A17B (thinking and non-thinking modes)~\cite{qwen3.5}, Gemini 3.1 Pro~\cite{geminipro}, Gemini 3 Flash~\cite{geminiflash3} and 3.5 Flash Lite~\cite{geminiflash35}, and GPT-5.5~\cite{gpt55} under zero-shot prompting. We additionally evaluate two top-performing models using few-shot in-context learning, with two demonstrations per interaction type. Each demonstration provides the ground-truth group members and joining pose together with the corresponding formation rule. All VLM baselines receive the same RGB image, depth map, and natural language instruction as our method, with persons annotated by unique indices. They predict (i) group member indices, (ii) an image-space goal position informed by depth, and (iii) a goal orientation as a top-down 2D unit vector under a shared axis convention. Since our method directly predicts metric top-down outputs, VLM image-space predictions are projected into the same coordinate frame using depth and estimated camera intrinsics. Predicted points landing on human bodies are snapped to the nearest free ground position (Fig.~\ref{fig:exp_result}); if none exists, the original prediction is retained. This post-processing favors the VLM baselines by correcting physically infeasible predictions rather than counting them as failures. Each VLM baseline runs $10$ trials per scenario. 

\textbf{Oracle-group baselines.}
To isolate goal-prediction performance from grounding errors, we provide the oracle-group and compare with a handcrafted baseline following interaction-specific geometric rules from the F-formation catalogue~\cite{barua2024enabling} and a multivariate regression approach~\cite{hedayati2023should} for conversational-group positioning. Following~\cite{hedayati2023should}, separate regression models are trained for different F-formation sizes on the same simulated data used by our model. These deterministic baselines produce one prediction per scenario.

\textbf{Real-robot baselines.}
We compare with two VLA navigation methods, NaVILA~\cite{cheng2024navila}, OmniVLA~\cite{hirose2025omnivla}, and a VLM-based sampling-and-selection baseline for real-robot evaluation. NaVILA is trained on large-scale web human-touring videos, some containing family-gathering scenarios, while OmniVLA is trained on SCAND~\cite{karnan2022socially}, which provides more than half of our test scenarios. The sampling-and-selection baseline uses the same top-down representation as our method, samples collision-free candidate poses, visualizes them with indices, and queries GPT-5.5 to select the most appropriate pose given the formation rules. Each method is evaluated over $5$ trials per scenario.

\textbf{Metrics.} 
For \textit{group grounding}, we conduct a human annotation study with $27$ participants, who identify target group members from scenes with labeled person indices. Inter-annotator agreement weights \textbf{precision}, \textbf{recall}, and \textbf{F1 score}. For \textit{goal-pose prediction}, we use three metrics derived from social conventions (e.g., joining a queue from the tail) and proxemics theory. 
\textbf{Position validity} requires a $0.5$--$1.2$m distance from relevant members: the last queue member, any conversation member around the group centroid, or any audience viewer while remaining behind the first row and not occluding others. \textbf{Orientation validity} requires a heading within $\pm30\degree$ of the queue direction, group centroid, or focal entity, respectively. \textbf{Pose validity} requires both. We report validity rates as mean $\pm$ standard deviation for oracle and predicted groups, along with inference time. For our method, $10$ goal poses are sampled per scenario, matching VLM trials. \textit{Real-robot} metrics are task success rate (\textbf{SR}; reaching a final pose that satisfies the same interaction-specific pose-validity criteria used offline), collision rate (\textbf{CR}), navigation time (\textbf{NT}), and personal-space violation duration (\textbf{PSV}), defined as the time spent within $0.25$m of any person~\cite{narasimhan2025olivia}.

\subsection{Experiment Results}

\textit{Group Grounding.}
Table~\ref{tab:ground_result} evaluates the first stage independently. Our method achieves an agreement-weighted F1 of $0.8125$ in $0.936$s, averaged over scenes containing $4$ to more than $20$ people. Although the strongest few-shot VLM achieves a slightly higher F1, it requires substantially longer inference time; other VLMs with comparable grounding performance are similarly slower. This accuracy--efficiency trade-off supports the intended role of our grounding module: providing a reliable target-group representation at a latency practical for downstream online goal prediction and updates.

\textit{Goal Prediction on Oracle Groups.}
The lower panel of Table~\ref{tab:pose_result} isolates goal prediction using oracle groups. Our method achieves high goal-pose validity across all interaction types. The handcrafted F-formation baseline remains sensitive to noisy poses and irregular group configurations, while the regression baseline struggles to model the multimodality of valid joining poses despite being trained on the same data. These results show that goal generation remains non-trivial even when group membership is known.

\begin{figure}[]
    \centering
    \includegraphics[width=0.98\linewidth]{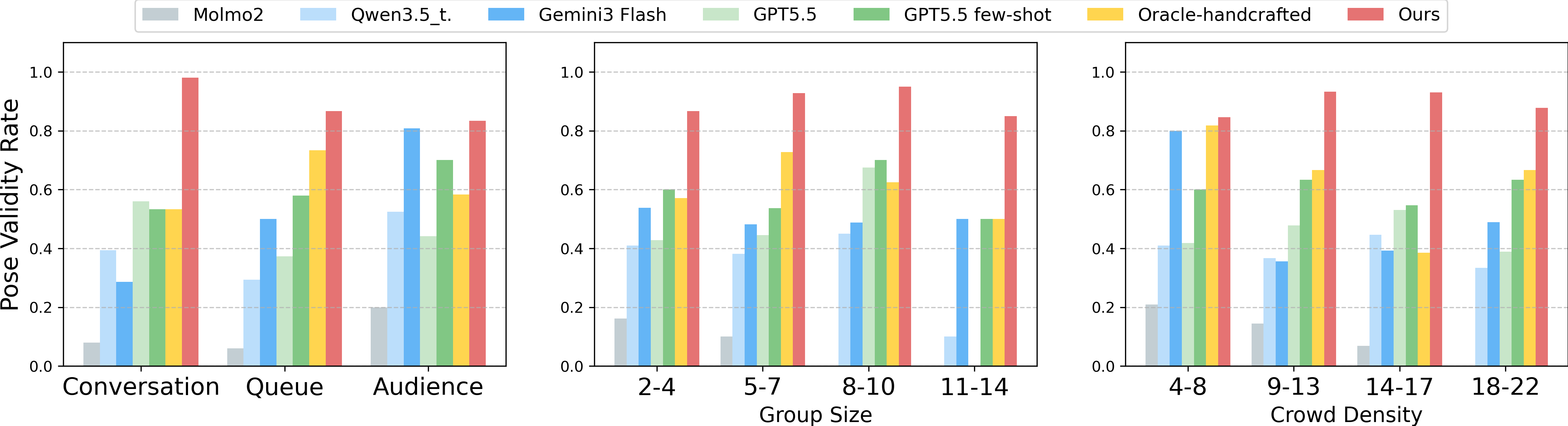}
    \vspace{-1em}
    \caption{Pose validity by interaction category, group size, and crowd density.}
    \label{fig:exp_pose_bycate}
    \vspace{-1.8em}
\end{figure}

\textit{End-to-End Offline Evaluation}. We first evaluate the complete pipeline on the offline benchmark, where goal prediction is conditioned on the group produced by the grounding stage, as shown in the upper panel of Table~\ref{tab:pose_result}. Our method achieves 89.76\% pose validity at approximately $1.1$ seconds per query. Few-shot in-context learning improves VLM pose validity but remains below ours, and its longer inference time limits applicability to dynamic scenes. Pose validity by interaction category, crowd density, and group size is further analyzed in Fig.~\ref{fig:exp_pose_bycate}. Gemini 3 Flash is comparatively competitive on audience scenes, likely because the visually salient focal entity simplifies both group identification and goal prediction. Comparing oracle-group and end-to-end results also reveals how grounding errors propagate to goal prediction. The effect is particularly pronounced for queues, where the identity of the terminal member directly determines the valid joining region. With oracle groups, the handcrafted method becomes more competitive in this setting, emphasizing the importance of accurate grounding.

Across VLM baselines, orientation validity is consistently higher than position validity. This is expected: orientation can often be inferred from visible body poses, whereas a valid joining position requires reasoning about metric distances in 3D space. Even with depth input, VLMs frequently fail to map image-space predictions to geometrically valid ground-plane coordinates. The qualitative results in Fig.~\ref{fig:exp_result} show that our energy--orientation maps capture valid, multi-modal goal distributions that adapt to group configurations while maintaining physically plausible orientations.

\textit{Real-Robot Evaluation}.
We deploy the system on a Boston Dynamics Spot legged robot with onboard NVIDIA Jetson Orin computation. Person detection is performed using YOLOv10~\cite{wang2024yolov10}, which is also used for person indexing in VLM inputs and our grounding module. Human orientation is estimated using~\cite{jiang2024rtmw}, which aligns with what we use for our goal prediction. Depth is obtained from Hesai JT128 LiDAR point clouds, and the robot is controlled via a DWA controller. In the tested medium-density scenes (i.e., $<$10 people), full inference takes $<$0.7s and runs continuously and asynchronously with control using globally tracked human positions. To avoid unnecessary goal changes, if group membership remains unchanged, the current goal is retained when it lies within a threshold ($0.4$m) of any of the top-5 newly predicted candidates; otherwise, it is replaced by the new lowest-energy prediction. When a new person joins a group, the goal is updated only after the membership change persists for three consecutive frames. Fig.~\ref{fig:robot_exp} shows snapshots of two test cases. Our algorithm successfully navigates the robot to the target group while handling appearing bystanders and updating goals based on new group members.

Table~\ref{tab:robot_result} shows the quantitative evaluation. As VLA methods have no explicit stopping criterion, navigation time is measured until the robot passes and moves away from the human. Our method succeeds in 18/20 trials, with the two failures caused by controller and detection errors. The sampling-and-selection baseline incurs longer inference delays and is less reliable in the dynamic queue. In two trials, it succeeds only because its initial goal lies more than $1.5$m behind the queue end and happens, by chance, to fall behind the newly joined person. This incidental behavior can also cause collisions. Its long inference time (shown in brackets, during which the robot remains stationary at the beginning on average) makes timely goal updates impractical.

\subsection{Ablation and Pilot Human Study}
We ablate the number of threshold-sweeping intervals used in hypothesis generation. As shown in Table~\ref{tab:abla_and_survey} (top), increasing the number of intervals improves grounding accuracy but also increases inference time; performance saturates near six intervals, which provides the accuracy--efficiency trade-off used in the main experiments. To complement rule-based validity metrics derived from the formation conventions in~\cite{barua2024enabling}, we compare the predicted joining-position distributions with independent human judgments from $15$ participants. Given an RGB image and instruction, participants click the position they consider most suitable for joining the group. We discretize each image into $500$ bins and compare the human distribution with model predictions using KL divergence. Table~\ref{tab:abla_and_survey} (bottom) shows that our method obtains the lowest divergence among the evaluated models, providing complementary evidence that the learned joining distribution reflects human judgments.

\begin{table}
\centering
\caption{Ablation and pilot human-study results}
\label{tab:abla_and_survey}
\vspace{-1em}
    \begin{minipage}{\linewidth}
    \centering
    \textbf{(a) Effect of sweeping intervals on group grounding}\\
    \setlength{\tabcolsep}{5pt}
    \resizebox{\linewidth}{!}{%
    \begin{tabular}{@{}lcccc@{}}
    \toprule
        \# Intervals & 8 & 6 & 4 & 2 \\
    \midrule
        F1/Time (s) & 0.8159 / 1.024 & 0.8125 / 0.936 & 0.7832 / 0.822 & 0.7202 / 0.635 \\
    \bottomrule
    \end{tabular}%
    }
    \end{minipage}
    
    \vspace{0.6em}
    
    \begin{minipage}{\linewidth}
    \centering
    \textbf{(b) KL divergence between predicted distributions and human survey}\\
    \setlength{\tabcolsep}{3pt}
    \resizebox{\linewidth}{!}{%
    \begin{tabular}{@{}l|c|c|c|c|c|c@{}}
    \toprule
        & Qwen3.5 (t.) & Gemini 3.1 Pro  & Gemini 3 Flash  & GPT-5.5  & GPT-5.5 few-shot       & \textbf{Ours}     \\
    \midrule
        KLD ($\downarrow$) & 20.40$\pm$1.29 & 18.99$\pm$2.75 & 17.75$\pm$3.61 & 19.60$\pm$1.79 & 19.51$\pm$1.07 & 14.87$\pm$1.08  \\
    \bottomrule
    \end{tabular}%
    }
    \end{minipage}
\vspace{-2.2em}
\end{table}

\section{Conclusion and Limitation}
\label{sec:conclusion}
We introduced language-grounded robot group joining, in which the navigation goal is inferred from a language-specified social interaction rather than given in advance. The approach first grounds the target interaction by ranking structured candidate groups and then predicts a multimodal joining-pose distribution using formation-derived attention priors. Experiments across diverse group sizes, crowd densities, and visually ambiguous scenes show that the method grounds language-specified groups at near-real-time speeds, while predicting valid joining poses across varied group types and spatial arrangements. Real-world experiments further demonstrate effective group joining.

\textbf{Failure Analysis and limitation.} We analyze five scenes with low or zero pose validity across 10 goal samples and identify two main causes: grounding errors and overly conservative joining distance. Since grounding relies on hypotheses generated from a distance-based affinity matrix, the correct group may be absent from the candidate subset (e.g., a customer standing at twice the queue spacing among others in Scene 21). The second failure mode occurs mainly in audience scenes, where the goal predictor assigns lowest energy to the outermost arc even when inner positions are available, leading to unnecessarily distant joining poses. Finally, a limitation of our approach is that the current formulation treats the environment mainly as a feasibility constraint rather than a direct source of joining affordances.


\section*{ACKNOWLEDGMENT}
Cited generative AI models were used for training-data generation and as experimental baselines. ChatGPT was used for language polishing and grammar correction. Authors are fully responsible for the content, analyses, and conclusions.

\end{document}